\documentclass[9pt,conference]{IEEEtran}
\IEEEoverridecommandlockouts % show footnote comments
\usepackage{blkarray}                                      % to support matrix
\usepackage{algpseudocode}                                 % new algorithm package
\usepackage{algorithm}
\usepackage{graphicx}                                      % include pdf figures
\usepackage{amsmath}
\usepackage{amssymb}
\usepackage{amsfonts}
\usepackage{amsthm}
\usepackage[mathcal]{eucal}
\usepackage{mathrsfs}
\usepackage{booktabs}
\usepackage{enumerate}
\usepackage{multirow}
\usepackage[subrefformat=parens,farskip=0pt,justification=centering]{subfig}
\usepackage{color}
\usepackage{cite}                                          % more citations in one bracket
\usepackage{comment}                                       % use comment
\usepackage{soul}                                          % use highlight command \hl{}
\soulregister\cite7
\soulregister\ref7
\soulregister\pageref7
\usepackage{etoolbox}                                      % commands \newtoggle, \toggletrue, \iftoggle
\usepackage{url}
\usepackage{nth}                                           % nth command
\usepackage{bm}                                            % bm command
\usepackage{courier}
\usepackage{balance}
\usepackage{threeparttable}
\usepackage{xcolor,colortbl}
\usepackage{footnote}

\usepackage{verbatim}
\usepackage[bookmarks=false]{hyperref}
\hypersetup{
    colorlinks = true,
    citecolor  = blue,
    linkcolor  = blue,
    urlcolor   = blue,
}
\usepackage{tikz}
\usetikzlibrary{patterns,snakes}
\usetikzlibrary{positioning,calc,fit,decorations.pathmorphing,shapes.geometric, shapes.gates.logic.US, calc}
\usetikzlibrary{arrows,arrows.meta,decorations.markings,shapes,shapes.arrows}
\usetikzlibrary{decorations,decorations.pathreplacing}
\usetikzlibrary{backgrounds}
\usepackage{filecontents}                                  % support to pgfplots
\usepackage{pgfplots}
\usepackage{pgfplotstable}
\usepackage{scalefnt}
\pgfplotsset{compat=newest}
\usepackage{caption}
\usepackage{cleveref}
\Crefformat{figure}{Fig.~#2#1#3}                           % "Fig.", instead of "Figure"
\Crefname{subfigure}{Fig.}{Figs.}
\Crefname{figure}{Fig.}{Figs.}
\Crefformat{table}{TABLE~#2#1#3}                           % "TABLE", instead of "Table"
\usepackage[figuresright]{rotating}
\definecolor{CUHKorange}{RGB}{244,106,18} %F47012
\definecolor{CUHKblue}{RGB}{0,111,190}    %006FBE
\definecolor{CUHKgreen}{RGB}{0,127,128}   %007F80
\definecolor{CUHKred}{RGB}{228,46,36}     %E42E24
\definecolor{CUHKyellow}{RGB}{198,148,34} %C69422
\definecolor{CUHKdark}{RGB}{114,44,114}   %722C72
\definecolor{CUHKmiddle}{RGB}{144,44,144} %902C90
\definecolor{CUHKlight}{RGB}{167,44,167} 
\definecolor{CUHKpurple}{RGB}{117,15,109}
\definecolor{CUHKgold}{RGB}{221,163,0}
\definecolor{CUHKribbon}{RGB}{244,223,176}
\definecolor{CUHKblack}{RGB}{34,24,21}

\newcommand{\tool}[1]{$\mathsf{#1}$}
\renewcommand{\bf}[1]{\textbf{#1}}
\renewcommand{\tt}[1]{\texttt{#1}}
\renewcommand{\vec}[1]{\boldsymbol{#1}}    % re-define vec command

\newcommand{\minisection}[1]{\vspace{.1in}\noindent{\textbf{#1}}}

\usepackage{tcolorbox}
\tcbuselibrary{skins,breakable}
    {\endtcolorbox}

\usepackage[top=0.75in,bottom=0.80in,left=0.58in,right=0.58in]{geometry}
\newtheorem{myproblem}{\textbf{Problem}}
\newtheorem{mydefinition}{\textbf{Definition}}

\crefname{mytheorem}{Theorem}{Theorems}
\crefname{mylemma}{Lemma}{Lemmas}
\crefname{myclaim}{Claim}{Claims}
\crefname{myproperty}{Property}{Properties}
\crefname{mycorollary}{Corollary}{Corollaries}

\algrenewcommand\textproc{\texttt}

\makeatletter
\let\OldStatex\Statex
\renewcommand{\Statex}[1][3]{%
  \setlength\@tempdima{\algorithmicindent}%
  \OldStatex\hskip\dimexpr#1\@tempdima\relax
}
\makeatother

\RequirePackage[normalem]{ulem} %DIF PREAMBLE
\RequirePackage{color}\definecolor{RED}{rgb}{1,0,0}\definecolor{BLUE}{rgb}{0,0,1} %DIF PREAMBLE
\usepackage{times}
\graphicspath{{../}{./figs/}}

\definecolor{myorange}{RGB}{238,97,42}  %
\definecolor{myblue}{RGB}{178,179,249}  %A2A0FE
\definecolor{mygrey}{RGB}{166,166,166}  %
\definecolor{mygreen}{RGB}{180,210,36}  %B4D224
\definecolor{myred}{RGB}{238,0,0}       %EE0000
\definecolor{myyellow}{RGB}{198,148,34} %C69422
\definecolor{mydark}{RGB}{114,44,114}   %722C72
\definecolor{mymiddle}{RGB}{144,44,144} %902C90
\definecolor{mylight}{RGB}{167,44,167}  %A72CA7
\definecolor{myblue1}{RGB}{137,157,192}  
\definecolor{mygreen1}{RGB}{69,137,148}

\begin{document}
\date{}

\title{
    %DiffPattern: Reliable Layout Pattern Generation with Discrete Diffusion Method
    \tool{DiffPattern}: Layout Pattern Generation via Discrete Diffusion
}

% \iffalse
\author{
    Zixiao Wang$^{1*}$ \thanks{* Equal contribution}\quad
    Yunheng Shen$^{2*}$, \quad
    Wenqian Zhao$^1$, \quad
    Yang Bai$^1$, \quad
    Guojin Chen$^1$, \quad
    Farzan Farnia$^1$, \quad
    Bei Yu$^1$ \\
    $^1$Chinese University of Hong Kong \quad $^2$ Tsinghua University  \\
    {\tt\small \{zxwang22, wqzhao, ybai, gjchen21, farnia, byu\}@cse.cuhk.edu.hk} \quad {\tt\small shenyh19@mails.tsinghua.edu.cn}
}
% \fi

\maketitle
\pagestyle{plain}

\begin{abstract}
% Reliable layout pattern libraries are the foundation of many lithography design works.
% Establishing practical Layout pattern libraries requires the ability to generate diverse patterns following specific design rules.
% Deep generative models dominate the existing literature in layout pattern generation. 
% However, leaving the guarantee of legality to an inexplicable neural network is not a sound choice. 
% In this paper, we propose DiffPattern to generate reliable layout patterns with three new strategies: 
% 1) A lossless layout pattern representation method dubbed as Deep Squish Pattern for efficient computation. 
% Deep Squish Pattern folds a squish pattern to a compact one with multiple channels to enlarge information density; 
% 2) A novel diverse deep topology generation method on the basis of the discrete diffusion model.
% We flip every pixel in a finite state space with estimated probability instead of thresholding on a continual output;
% 3) A white-box 2D pattern assessment method to transfer a deep topology to a legal pattern given desired design rules. 
% Experiments on the standard benchmark show that DiffPattern significantly outperforms existing baselines and is able to synthesize reliable layout patterns.

% Establishing practical Layout pattern libraries requires the ability to generate diverse patterns following specific design rules.
    Deep generative models dominate the existing literature in layout pattern generation. 
    However, leaving the guarantee of legality to an inexplicable neural network could be problematic in several applications. 
    % In this paper, we propose \tool{DiffPattern} to generate reliable layout patterns with three important properties: 
    % 1) A novel diverse deep topology generation via a discrete diffusion model.
    % 2) A lossless layout pattern representation for efficient computation. 
    % 3) A white-box pattern assessment to generate legal patterns given desired design rules. 
    % Our experiments on several benchmark settings show that \tool{DiffPattern} significantly outperforms existing baselines and is capable of synthesizing reliable layout patterns.
    In this paper, we propose \tool{DiffPattern} to generate reliable layout patterns. \tool{DiffPattern} introduces a novel diverse topology generation method via a discrete diffusion model with compute-efficiently lossless layout pattern representation. Then a white-box pattern assessment is utilized to generate legal patterns given desired design rules. Our experiments on several benchmark settings show that \tool{DiffPattern} significantly outperforms existing baselines and is capable of synthesizing reliable layout patterns.

\end{abstract}

%%%%%%%
%\begin{IEEEkeywords}

%\end{IEEEkeywords}

\section{Introduction}
\label{sec:intro}   

Reliable Very-Large-Scale Integration (VLSI) layout pattern libraries are the foundation of various designs for manufacturability (DFM) research, including perfection of design rules, Optical Proximity Correction (OPC) recipes \cite{gao2014mosaic}, lithography simulation \cite{kuang2013efficient,yu2015layout}, layout hotspot detection \cite{chen2019faster}, and so on. With the rapidly growing requirement for layout patterns in machine-learning-based lithography design applications, building a practical large-scale pattern library could be highly time-consuming due to the long logic-to-chip design cycle. 

Various rule-based and learning-based layout pattern generation methods have been proposed to address this issue.
Early rule-based methods \cite{reddy2018enhanced, ye2019lithoroc} first augment a set of predefined basic units by simple enhancement technology, {\it e.g.,} flipping and rotation. Then they randomly select several units and splice them together. However, patterns generated in this way lack diversity and are limited in quantity. Recently, learning-based generative methods \cite{yang2019deepattern, zhang2020layout, wen2022layoutransformer} have shown the ability to produce a large number of diverse layout patterns. Among them, pixel-based methods \cite{yang2019deepattern, zhang2020layout} regard the pattern generation problem as a binary image generation task.
A lossless pattern representation, {\it Squish Pattern}, is proposed by \cite{gennari2014topology} to reduce computational costs. Squish Pattern splits a large layout pattern into a significantly smaller binary topology matrix and two geometric vectors. Pixel-based methods synthesize a topology matrix with continual values and threshold it to get a new binary topology matrix, which wastes computation and hurts model capacity. Also, the sequential-based method 
in \cite{wen2022layoutransformer} models a layout pattern as a sequence of polygons, which is further decomposed into a sequence of vertices and directed edges. To get a new layout pattern, \cite{wen2022layoutransformer} generates a polygon sequence and translates it into a layout pattern. For both kinds of them, the generative model learns a latent regularization from the training set to avoid producing illegal layout patterns that violate the design rules.
We argue that the implicit constraint learned from the training set may be neither flexible nor reliable. Apart from the inconvenience that we need to train a new model on a specific large-scale dataset that follows the new design rules when the design rule changes, a considerable ratio of generated patterns violate the design rules in these methods.

In this paper, we propose \tool{DiffPattern}, a pixel-based practical layout pattern generation framework that consists of three main components: a) Inspired by the great success of diffusion models \cite{ho2020denoising,song2020denoising}, we approach the generation of topology as a denoising task for a random noise image. We utilize a discrete diffusion model to predict the noise that should be removed at every step. During the training and sampling procedure, each entry of the image tensor can be expressed by a discrete state in a finite state space, $\{0,1\}$ for example, and there is no need to manually set a threshold on a continual output range as previous work does. The naturally discrete output provides a reasonable regularization that avoids meaningless overfitting on how to produce a discrete-like output since all the pixels in the training samples are discrete in their values and position.
b) To further enlarge the information density of patterns and reduce computation costs, \tool{DiffPattern} adapts the idea of Squish Pattern Representation and pushes it one step forward. A kind of lossless layout pattern representation method called {\it Deep Squish Pattern} is proposed. Deep Squish Pattern folds a topology matrix into a topology tensor, which shrinks the input size and extends the channel dimension. Due to the fact that the efficiency of the existing diffusion model is more sensitive to the input size \cite{yang2022diffusion} and significantly less to the number of input channels, Deep Squish Pattern provides an easy-to-compute solution and can be efficiently applied to other pixel-based pattern generation methods. c) With newly generated topology matrices, we aim to assign geometric vectors to them and restore legal layout patterns. We develop a nonlinear system that can figure out the legal solution for each topology matrix and can be easily adjusted to every design rule.
Empowered by the interpretable pattern assessment strategy, \tool{DiffPattern} achieves a notable 100\% legality rate on generated layout patterns in our setting.

% In this subsection, we will introduce the detailed generation method of squish patterns based on the diffusion model. Diffusion models usually refer to denoising diffusion probabilistic models (DDPM), which have recently shown great potential in generating high-quality image samples, even outperforming previous VAEs, GANs, and Flow-based models. Due to its unique idea of adding noise and denoising, the diffusion models keep latent variables of the same dimension as the input rather than embedding them in a lower dimension, which also makes the diffusion models have a larger feature representation space than the other generative models mentioned above. In addition, unlike VAEs that require a trade-off between generative diversity and generative similarity, GANs whose training process is unstable due to adversarial training, and Flow-based models that require building complex reversible transformations and specific architectures, the diffusion models can directly use general neural network architectures, have a stable training loss function, and ensure the diversity and similarity of the generated samples at the same time.

% The diffusion models considers the process of adding noise and denoising to the sample, and represents it as a Markov chain. They slowly added random noise to the data in the forward diffusion process, and then learned the corresponding reverse diffusion process through the neural network. The reverse diffusion process can be used to generate data samples from random noise that conform to the original distribution.

Our main contributions can be summarized as follows:
\begin{enumerate}
    \item We develop a novel layout pattern generation method based on discrete denoising for synthesizing layout topology.
    \item We propose a lossless layout pattern representation strategy, Deep Squish Pattern, which accelerates pixel-based layout pattern generation schemes.  
    \item We utilize a nonlinear system for pattern assessment where the system provides a white-box method to legalize the layout patterns.  
    \item We extensively evaluate our methodology on benchmark datasets showing that \tool{DiffPattern} can achieve state-of-the-art (SOTA) performance.
\end{enumerate}

\section{Preliminaries}

\begin{figure}[tb!]
    \centering
    \includegraphics[width=0.76\linewidth]{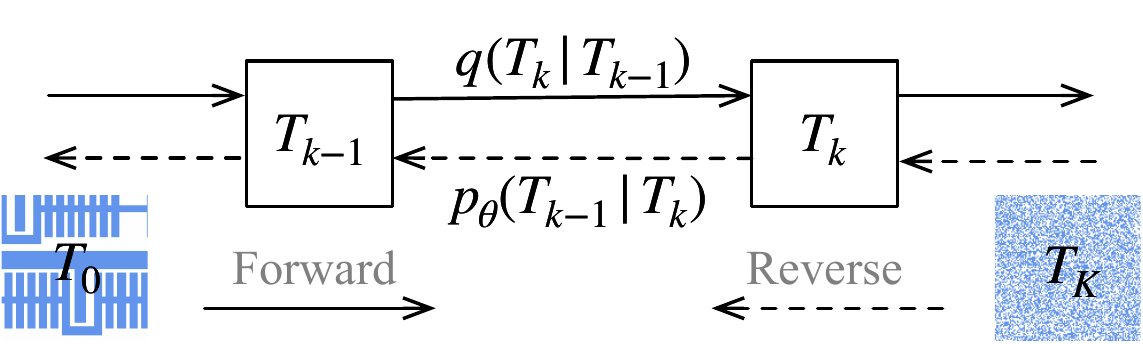}
    \caption{Illustration of denoising diffusion process. }
    \label{fig:ddpm}
\end{figure}

\subsection{Diffusion Models}
\label{sec:2.1}

Diffusion models refer to denoising diffusion probabilistic models \cite{ho2020denoising}, which have  demonstrated great potential in generating high-quality images. A diffusion model defines a Markov chain to represent the sample forward and reverse diffusion process, as shown in \Cref{fig:ddpm}. In the forward diffusion process, the diffusion model produces a sequence of noisy samples $\vec{T}_1,...,\vec{T}_K$ by iteratively adding Gaussian noise to the original sample $\vec{T}_0$ in $K$ steps, where the noise level is controlled by a variance schedule $\left\{ \beta_k \in (0,1) \right\}_{k=1}^K$:
\begin{equation}
    \vec{q}\left(\vec{T}_k | \vec{T}_{k-1}\right) := \mathcal{N}\left(\vec{T}_k ; \sqrt{1-\beta_k} \vec{T}_{k-1}, \beta_k \vec{I}\right).
    \label{eq:ddpm-forward}
\end{equation}

When $K$ is large, the noisy sample at the final iteration $\vec{T}_K$ will have an almost Gaussian distribution, which inspires the diffusion model to construct a reverse diffusion process to produce fresh data samples from randomly drawn Gaussian noise. The goal of the reverse diffusion process is to learn the  distribution of inverting the explained forward diffusion process, which will be used to generate fresh data samples from a Gaussian noise input. However, a proper inversion of the distribution requires  an expressive model, and therefore we use a deep neural network with learnable parameters $\vec{\theta}$ to learn the distribution,
% \begin{equation}
% p_\theta\left(\vec{x}_{0: T}\right)=p\left(\vec{T}_k\right) \prod_{t=1}^T p_\theta\left(\vec{T}_{k-1} \mid \vec{T}_k\right)
% \end{equation}
\begin{equation}
    \vec{p}_{\vec{\theta}} \left(\vec{T}_{k-1} | \vec{T}_k\right) := \mathcal{N}\left(\vec{T}_{k-1} ; \vec{\mu}_{\vec{\theta}} \left(\vec{T}_k, k\right), \vec{\Sigma}_{\vec{\theta}} \left(\vec{T}_k, k\right)\right).
    \label{eq:ddpm-reverse}
\end{equation}

Similar to variational autoencoders (VAEs), the training of diffusion models aims to maximize the log-likelihood function by optimizing the variational lower bound.
\begin{equation}
    L_{\mathrm{VLB}} = D_{\mathrm{KL}}\left(\vec{q}\left(\vec{T}_K | \vec{T}_0\right) \parallel \vec{p}_{\vec{\theta}}\left(\vec{T}_K\right)\right) + \sum_{k=2}^{K} L_{k} -\log \vec{p}_{\vec{\theta}}\left(\vec{T}_0 | \vec{T}_1\right),
    \label{eq:ddpm-loss}
\end{equation}
where $ L_k = D_{\mathrm{KL}}\left(\vec{q}\left(\vec{T}_{k-1} | \vec{T}_{k}, \vec{T}_0\right) \parallel \vec{p}_\theta\left(\vec{T}_{k-1} | \vec{T}_{k}\right)\right)$, $D_{\mathrm{KL}}$ is the KL divergence, and the term $\vec{q}\left(\vec{T}_{k-1} | \vec{T}_k, \vec{T}_0\right)$ can be derived a closed form of Gaussian distribution according to \Cref{eq:ddpm-forward} and Bayes' theorem.
% \begin{equation}
% \begin{aligned}
% q\left(\vec{T}_{k-1} | \vec{T}_k, \vec{T}_0\right)&=\frac{q\left(\vec{T}_k | \vec{T}_{k-1}, \vec{T}_0\right) q\left(\vec{T}_{k-1} | \vec{T}_0\right)}{q\left(\vec{T}_k | \vec{T}_0\right)}
% \end{aligned}
% \end{equation}

After training the diffusion model, we can simply sample from a standard Gaussian distribution and remove the noise in an iterative fashion according to \Cref{eq:ddpm-reverse} that follows the reverse diffusion process, resulting in new samples that follow the original sample distribution.

\subsection{Squish Pattern Representation}
\label{subsection:quish}

\begin{figure}[tb!]
    \centering
    \includegraphics[width=0.92\linewidth]{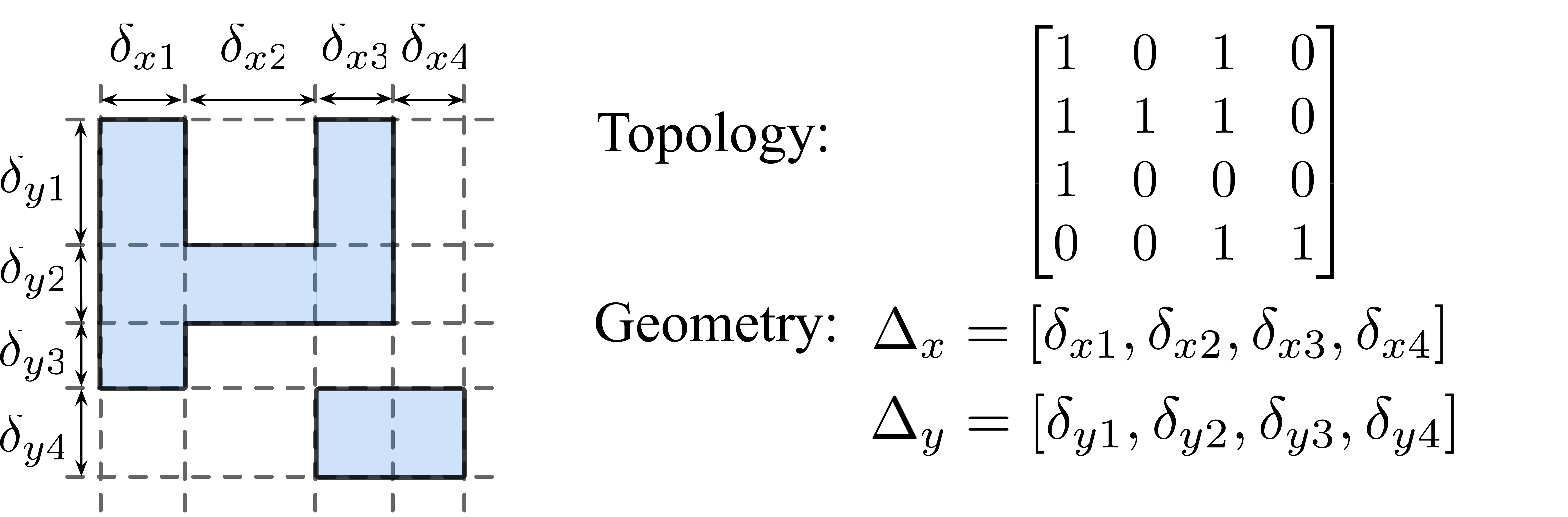}
    \caption{Squish Pattern Representation.}
    \label{fig:squishpattern}
\end{figure}

A typical layout pattern consists of a stack of polygons and is information-sparse, which leads to unnecessary computational costs and additional overfitting risk for neural network methods. The squish pattern \cite{gennari2014topology} is a lossless and efficient representation method which encodes a layout into pattern topology matrix and geometric information $\Delta_x,\, \Delta_y$, as shown in \Cref{fig:squishpattern}. The layout is split into grids by a set of scan lines that walk along the edges of polygons. The interval length of every adjacent scan line pair is stored in the $\Delta$ vectors. Every entry of the topology matrix is either zero or one, which indicates shape or zeros respectively. Every squish pattern is extended to a square with fixed side length by the method introduced in \cite{yang2019detecting}.
%To push model acceleration a step ahead, we proposed Deep Squish Pattern Representation, which has larger information density and is friendly to neural network. We will detail it in \Cref{subsec:compact}.

\subsection{Problem Formulation}
\label{sec:2.3}

%\noindent \textbf{Pattern Diversity}.
The diversity of generated patterns is a critical metric for the layout pattern generation task. As introduced in \cite{yang2019deepattern}, the complexity of a layout pattern is defined as $(c_x, c_y)$, where $c_x$ and $c_y$ are the numbers of scan lines subtracted by one along the x-axis and y-axis, respectively. Then we have,

%\noindent {\bf Pattern Diversity}
\begin{mydefinition}
The diversity of the patterns library, denoted by $H$, is defined as the Shannon Entropy of the distribution of the pattern complexities as follows:
\begin{equation}
    H = - \sum_i \sum_j P(c_{xi}, c_{yj})\log{P(c_{xi}, c_{yj})},
    \label{eq:diveristy}
\end{equation}
where $P(c_{xi}, c_{yj})$ is the probability of a pattern with complexity $(c_{xi}, c_{yj})$ sampled from the library.
\end{mydefinition}

In a typical case, a greater pattern diversity $H$ indicates that the library contains more widely distributed patterns.

\begin{figure}
    \centering
    \includegraphics[width=0.68\linewidth]{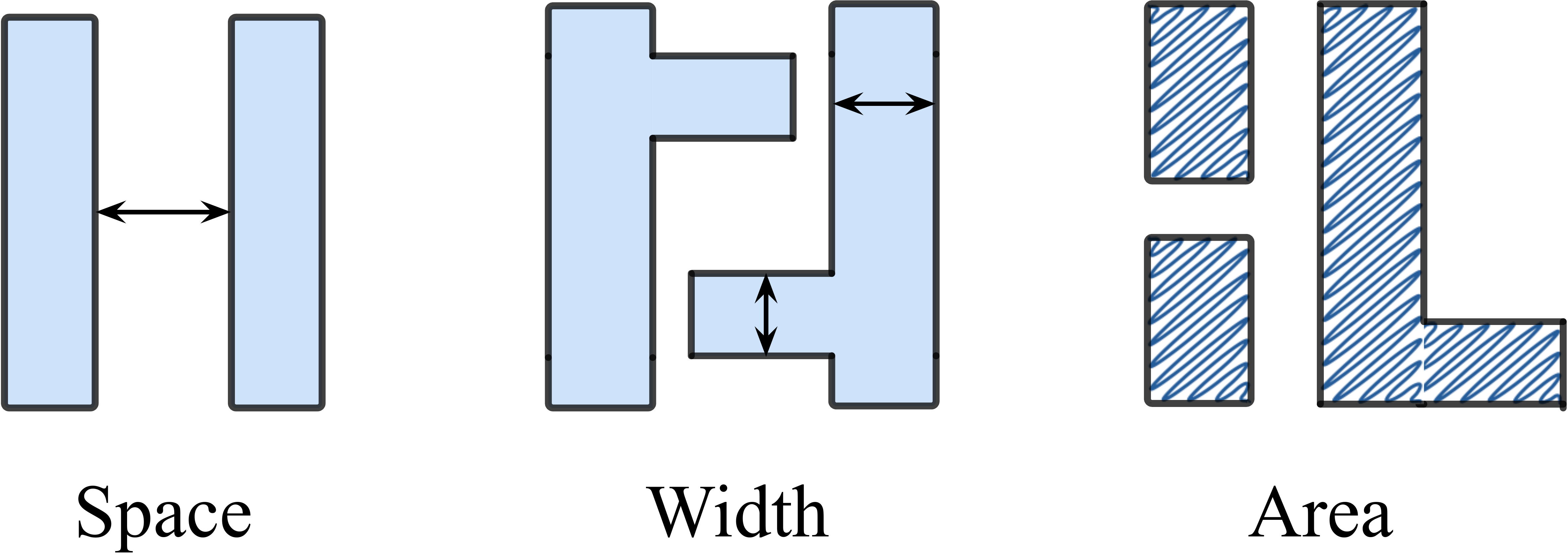}
    \caption{Illustration of design rules.}
    \label{fig:drc_rule}
\end{figure}

%\textbf{Pattern Legality}.
Based on the related works \cite{zhang2020layout,wen2022layoutransformer}, the generated patterns should satisfy several pre-defined design rules of IC layout. As illustrated in \Cref{fig:drc_rule}, `Space' represents the distance between two adjacent polygons. `Width' measures the size of a shape in one direction. And `Area' denotes the area of a polygon. Based on these geometric measurements, we have,

\begin{mydefinition}[Pattern Legality]
We call a layout pattern {\it legal} if the layout pattern is DRC-clean, given the design rules.
\end{mydefinition}

%\noindent {\bf Pattern Validity}.
% When we aim to synthesis realistic patterns, we hope the generated patterns share similar local features with existing layout patterns. To evaluate how realistic the generated patterns are, \cite{zhang2020layout} defines pattern validity as,

% \begin{mydefinition}{Pattern Validity}
%     Pattern validity is the ratio of realistic patterns to total generated patterns. 
% \end{mydefinition}

% In \Cref{subsection:validity}, we gives the scheme to measure the validity of the given pattern set, and we further discuss the limitation of validity. 
Based on the above evaluation metrics, the pattern generation problem can be formulated as follows,

\begin{myproblem}[Pattern Generation]
    Given a set of design rules and existing patterns, the objective of pattern generation is to synthesize a legal pattern library such that the pattern diversity of the layout patterns in the library are maximized.
\end{myproblem}

% \newpage
\section{Framework}

\begin{figure*}[th!]
    \centering
    \includegraphics[width=.88\linewidth]{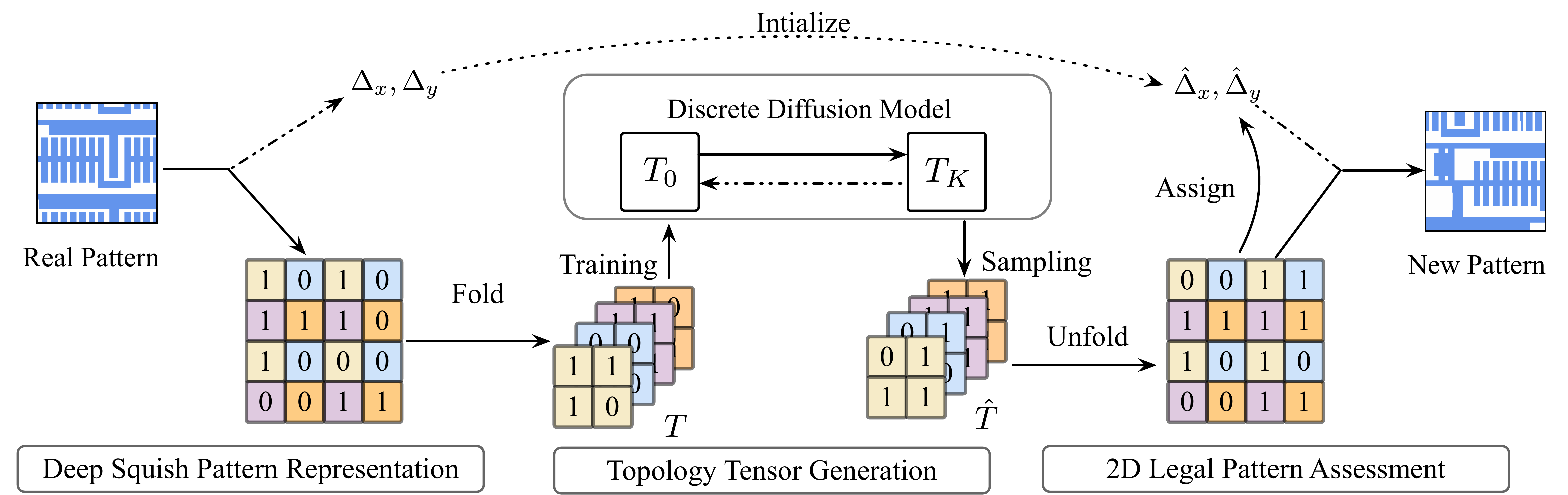}
    \caption{An illustration of the \tool{Diffpattern} framework for reliable layout pattern generation.  }
    \label{fig:pipeline}
\end{figure*}

\subsection{Overview of \tool{DiffPattern}}
As illustrated in \Cref{fig:pipeline}, our framework consists of three phases:
(1) {Deep Squish Pattern Representation.} Given a set of layout patterns, we first extract their deep squish pattern representation in which every layout pattern is decomposed into a topology tensor $\vec{T}$ and two geometric vectors $\vec{\Delta}_x$ and $\vec{\Delta}_y$.
(2) {Topology Tensor Generation.} We subsequently feed the extracted topology tensors $\vec{T}_0$ into a diffusion model. By gradually adding noise to $\vec{T}$ with fixed probabilities $q(\vec{T}_k|\vec{T}_{k-1})$, the model learns how to reverse this $K$-step process.
To synthesize a new topology tensor $\hat{\vec{T}}$, we randomly sampled a noise topology $\vec{T}_K$, and every entry in $\vec{T}$ jumps between finite states with estimated probability $p_\theta(\vec{T}_{k-1}|\vec{T}_k)$ until getting a reasonable topology tensor $\vec{\hat{T}}_0$. 
(3) {2D Legal Pattern Assessment.} Given the generated topology tensors $\vec{\hat{T}}$, we established an explainable nonlinear system to figure out the proper geometric vectors for each topology tensor according to the {\it Design Rules}.

\subsection{Deep Squish Pattern Representation}
\label{subsec:compact}

\begin{figure}[h]
    \centering
    \includegraphics[width=0.88\linewidth]{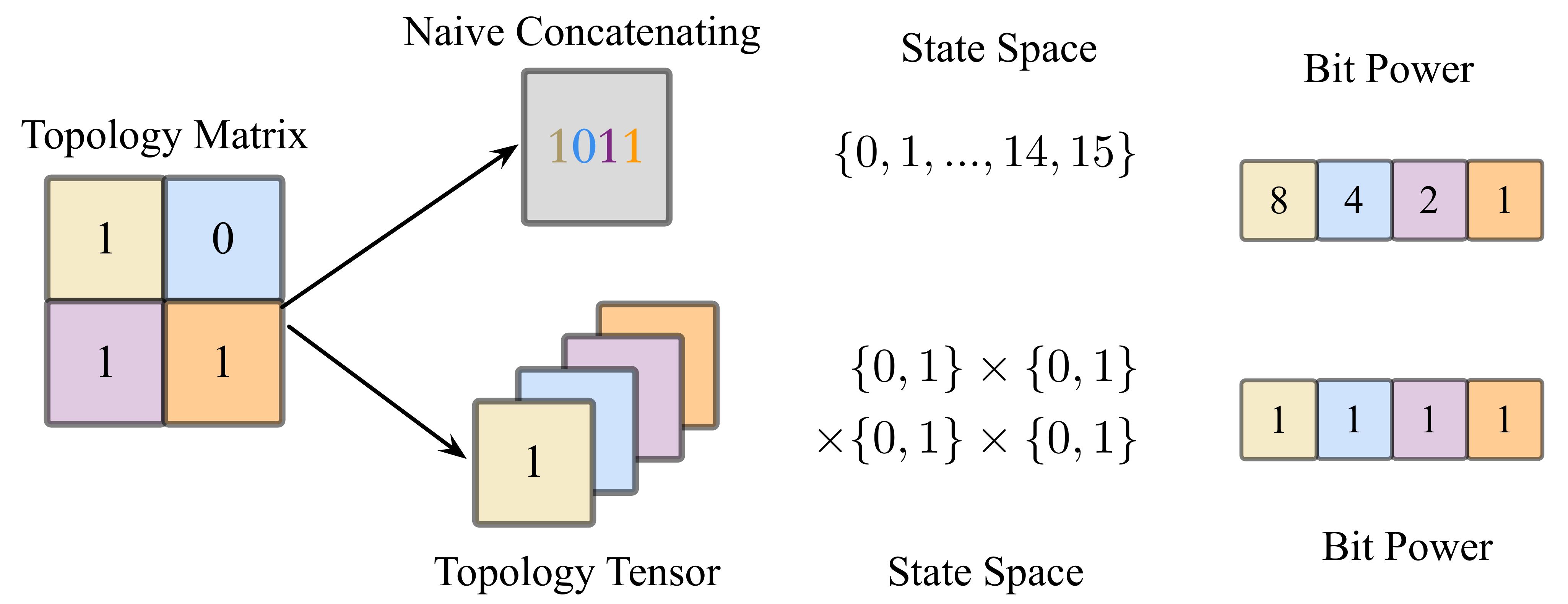}
    \caption{An illustration of Deep Squish Pattern Representation. The Topology Tensor is a lossless and compact representation of the topology matrix. And the Naive Concatenating brings unbalanced power to each bit and an exponentially increasing state space. }
    \label{fig:compact}
\end{figure}

As we have discussed in \Cref{subsection:quish}, the squish pattern is a lossless representation of the layout pattern, and the topology matrix can be treated as a one-channel 2D binary mask, as shown in \Cref{fig:compact} (left). However, the information density of each pixel is still not satisfactory given that the training/inference efficiency of diffusion methods is more sensitive to the size of input images but significantly less on the number of states of each pixel. To address this issue, we propose a novel pattern representation method, {\it Deep Squish Pattern,} to attain a more compact pattern representation.

Here we use a running example, as shown in \Cref{fig:compact}. There is a topology matrix with four (2$\times$2) adjacent pixels. Each pixel is assigned with either zero or one to indicate shape or space, respectively.
A simple idea to enlarge the information density is to encode the bits from multiple pixels into one pixel. However, directly concatenating every bit from different pixels and forming a new state number, {\it e.g.,} using 0-15 to represent 16 different states of (2$\times$2) adjacent pixels, will bring unbalanced power to each position and may import numerical instability to the network when the bit count is large. For example in the 4$\times$4 case, the first bit gets a power of $2^{15}$ while the last bit only gets a power of 1. Furthermore, the count of states increases exponentially with the count of bits.
% In the running examples, the green pixel will get a power of 8 while the orange gets a power of 1 in this case. 

% \yh{On the other hand, when the size of the selected folded patch increases, its logical representation state increases exponentially, for example, the number of states required by a $4\times4$ patch is $2^{4*4}=65536$, which makes the size of the transition probability matrix for these states even more terrifying when we use a discrete method.}

Since a state space with 16 ($2^4$) discrete states can be expressed by a permutation of four subspaces with two states. Instead of encoding different positions with different power, we assign equivalent weight to each position by folding the squish topology matrix into a topology tensor $\vec{T}$ with multiple channels. In the folding step, a patch with the size of $\sqrt{C}\times \sqrt{C}$ from the topology matrix is transferred to a point with $C$ channels in the tensor $\vec{T}$. $C$ is a hyperparameter, which is chosen by considering the trade-off between local information density and the input size. Deep Squish Pattern representation enlarges the practical receptive field of the model and is naturally suitable for pixel-based machine learning methods. Once the generation procedure ends, we can also equivalently recover the original topology matrix by flattening the topology tensor.
% \yh{$\vec{Q}=\vec{Q}^{bit3}\otimes\vec{Q}^{bit2}\otimes\vec{Q}^{bit1}\otimes\vec{Q}^{bit0}$}
%Therefore, we squeeze the squish topology representation on a small space (n*n) to the channel, so that different pixels have the same power. The hyperparameter n is chosen according to the size of the image to trade off information density and global spatial information, where we choose n=4 in this article. In fact, since the convolution kernels do not share parameters across channels, our representation is equivalent to using a larger kernel size and a larger stride in the first convolutional layer, which helps to get a larger receptive field, increase the information density and reduce the image size.

% Based on the assumption that the state transition distributions on each pixel are independent, a 16$\times$16 transition matrix can be expressed as the Kronecker product of four 2$\times$2 transition matrices. Furthermore, given the assumption of independent and identical distribution, the four transition matrices are the same matrix, that is, the transition matrix of 16$\times$16 can be completely derived from a matrix of 2$\times$2 in essence.

\subsection{Topology Tensor Generation}

Once we have encoded the existing layout pattern efficiently by using deep squish pattern representation, we aim to learn the distribution of existing topology tensors and generate new ones with reasonable topology attributes. Let ${\vec{T}_0}\in \{0,1\}^{C\times M\times M}$ be a topology tensor extracted from existing patterns. The naive idea is that treat the binary tensor as a grayscale image, learn the distribution through a diffusion model introduced in \Cref{sec:2.1} and clip the generated topology into a binary one by setting a threshold on it as previous pixel-based pattern generation methods \cite{zhang2020layout,yang2019deepattern} do. The problem of learning how to generate discrete values (zero and one in our case) at every entry is left to the network. However, we argue that requesting the network learn a discrete output in a continual state space from the training set is a waste of model representation ability. A more elegant way is to generate discrete output naturally. 

\minisection{Discrete Diffusion Model.}
% Therefore, we introduce a discrete diffusion model \cite{austin2021structured} to the binary topology generation problem. 
Unlike the traditional diffusion model utilized in the computer vision domain, we aim to synthesize the topology of layout pattern, where every entry in the topology belongs to a discrete state. We have made several key modifications to enhance the diffusion model and synthesize discrete topology patterns directly. To derive them, we first re-formulate the problem.

At the $k$-th of $K$ diffusion step, $x_k\in \{0,1\}$ is an entry in the topology tensor $\vec{T}$. In the discrete diffusion model, a transition probability matrix $[\vec{Q}_k]_{ij} = q(x_k=j|x_{k-1}=i)$ is defined to describe the state transition probability for each $\vec{x}$ at the $k$-th diffusion step,
\begin{equation}
\vec{q}\left(\vec{x}_k \mid \vec{x}_{k-1}\right) := \operatorname{Cat}\left(\vec{x}_k ; \vec{p}=\vec{x}_{k-1} \vec{Q}_k\right),
\label{eq:d3pm-forward}
\end{equation}
where $\vec{x}_k$ is the one-hot version of the entry $x_k$, $\operatorname{Cat}(\vec{x}|\vec{p})$ is a categorical distribution over the row vector $\vec{x}$ with probabilities given by the row vector $\vec{p}$, and $\vec{x}_{k-1} \vec{Q}_k$ can be understood as a row vector-matrix product. 
The $\vec{Q}_k$ is applied to each entry in the topology tensor independently and $\vec{q}$ factorizes over these higher dimensions as well. The proposed deep squish pattern representation is customized for this discrete diffusion process, since the size of transition matrix $\vec{Q}$ increases with the count of states of each pixel. Every entry $x$ in the topology tensor owns only two states. On the other side, in the reverse diffusion process, the neural network aims to predict the categorical distribution probability $\vec{p}_\theta(\vec{x}_{k-1}|\vec{x}_k)$ over each entry to recover the original tensor.
 
The choice of transition probability matrix $\vec{Q}_k$ is critical, which should ensure the forward process $\vec{q}(\vec{x}_k|\vec{x}_0)$ converging to a known stationary distribution when $k$ becomes large. A uniform stationary distribution is a natural choice in topology tensor generation, which means given any $\vec{x}_0$, the distribution of every entry $\vec{x}_k$ should follows,
\begin{equation}
    \vec{q}(\vec{x}_k|\vec{x}_0)\rightarrow \left[0.5,0.5\right],~\text{when}~k\rightarrow K.
\end{equation}
So we design a doubly stochastic matrix $\vec{Q}_k$ with strictly positive entries for topology denoising diffusion process,
\begin{equation}
\vec{Q}_k = \begin{bmatrix} 1-\beta_k & \beta_k \\ \beta_k & 1-\beta_k \end{bmatrix},
\end{equation}
where ${\beta_k} \in (0,1)$ is the hyperparameter controlling the noise level. In order to ensure that the model can learn the original sample distribution more finely and quickly reach a stable distribution, we follow the classical setting in previous works \cite{ho2020denoising,austin2021structured}. We set a smaller noise in the early diffusion step and a larger noise in the later step. Specifically, we use a linearly increasing schedule for $\beta_k$:
\begin{equation}
\beta_k = \frac{(k-1)\left(\beta_K-\beta_1\right)}{K-1}+\beta_1,~k = 1,...,K,
\end{equation}
where $\beta_1$ and $\beta_K$ are hyperparameters. 
\minisection{Training Diffusion Model.} To training the discrete diffusion model for topology tensor generation, the training objective at step $k$ is to minimize the loss function,
% We follow the loss function adopted by D3PM \cite{austin2021structured} which adds an auxiliary denoising loss on baisc of \Cref{eq:ddpm-loss}. In the actual training process, the above loss function can simplify and focus on a specific step $k$, as shown in \Cref{eq:d3pm-loss-true}.
% \begin{equation}
% L_{total}=L_{\mathrm{VLB}}+\lambda \mathbb{E}_{q\left(\vec{T}_0\right)} \mathbb{E}_{q\left(\vec{T}_k | \vec{T}_0\right)}\left[-\log p_{\vec{\theta}} \left(\vec{T}_0 | \vec{T}_k\right)\right]
% \label{eq:d3pm-loss}
% \end{equation}
\begin{equation}
L = D_{\mathrm{KL}}\left(\vec{q}\left(\vec{x}_{k-1} | \vec{x}_{k}, \vec{x}_0\right) \parallel \vec{p}_\theta\left(\vec{x}_{k-1} | \vec{x}_{k}\right)\right) - \lambda\log \vec{p}_{\vec{\theta}} \left(\vec{x}_0 | \vec{x}_k\right),
\label{eq:d3pm-loss-true}
\end{equation}
where $\lambda$ is a hyperparameter to balance the loss terms.

Given a topology tensor $\vec{T}_0$, we randomly sample a target step $k$ from $1$ to $K$ firstly, and expect to get the noisy sample $\vec{T}_k$. Fortunately, we can explicitly derive that $\vec{x}_k$ obeys the following categorical distribution:
\begin{equation}
    \vec{q}\left(\vec{x}_k|\vec{x}_0\right) = \operatorname{Cat}\left(\vec{x}_k ; \vec{p}=\vec{x}_0 \overline{\vec{Q}}_k\right),
\end{equation}
where $\overline{\vec{Q}}_k=\vec{Q}_1 \vec{Q}_2 \ldots \vec{Q}_k$.
% $q\left(\vec{T}_k|\vec{T}_0\right) = \operatorname{Cat}\left(\vec{T}_k ; \vec{p}=\vec{T}_0 \overline{\vec{Q}}_k\right)$, where $\overline{\vec{Q}}_k=\vec{Q}_1 \vec{Q}_2 \ldots \vec{Q}_k$
Then, we can directly sample from the above distribution to obtain $\vec{T}_k$, instead of adding noise $k$ times.

After sampling $\vec{T}_k$, we feed it into the neural network with the embedding of the time step $k$. The neural network will predict the logits of the posterior distribution $\vec{p}_{\vec{\theta}}\left(\vec{x}_0|\vec{x}_k\right)$, and then $\vec{p}_{\vec{\theta}}\left(\vec{x}_{k-1} | \vec{x}_k\right)$ can be calculated as following:
\begin{equation}
\vec{p}_{\vec{\theta}}\left(\vec{x}_{k-1} | \vec{x}_k\right) = \sum_{\widetilde{\vec{x}}_0} \vec{q}\left(\vec{x}_{k-1} | \vec{x}_k , \widetilde{\vec{x}}_0\right) \vec{p}_{\vec{\theta}}\left(\widetilde{\vec{x}}_0 | \vec{x}_k\right),
\label{eq:d3pm-reverse}
\end{equation}
where the term $\widetilde{\vec{x}}_0$ will visit every possible state of $\vec{x}_0$. And $\vec{q}\left(\vec{x}_{k-1} | \vec{x}_k, \widetilde{\vec{x}}_0\right)$ has a closed form according to \Cref{eq:d3pm-forward} and Bayes' theorem,
\begin{equation}
\vec{q}\left(\vec{x}_{k-1} | \vec{x}_k, \vec{x}_0\right) = \operatorname{Cat}\left(\vec{x}_{k-1} ; \vec{p}=\frac{\vec{x}_k \vec{Q}_k^{\top} \odot \vec{x}_0 \overline{\vec{Q}}_{k-1}}{\vec{x}_0 \overline{\vec{Q}}_k \vec{x}_k^{\top}}\right),
\label{eq:d3pm-closedform}
\end{equation}
where $\odot$ is a pixel-wise multiplication.

So far, all items in the loss function have been obtained, and the diffusion model can be trained by the commonly used gradient descent method.

\begin{figure}[tb!]
    \centering
    \includegraphics[width=0.88\linewidth]{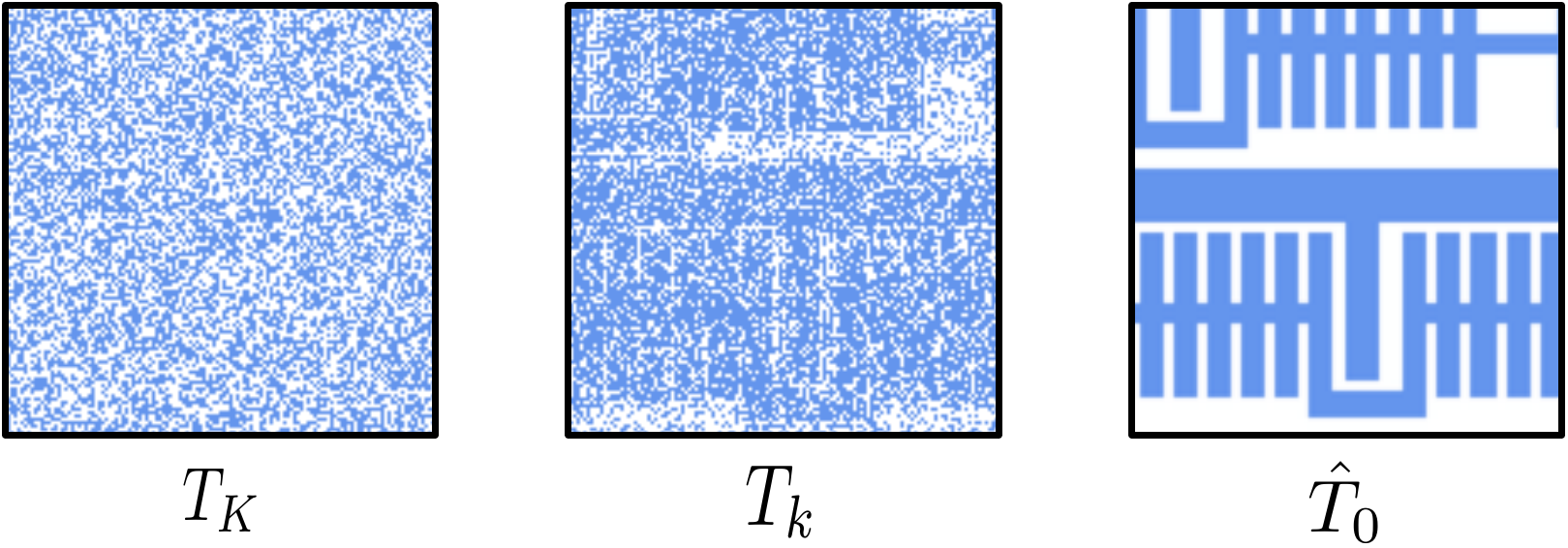}
    \caption{An illustration of the (flattened) samples from our Discrete Diffusion Model.}
    \label{fig:sampling}
\end{figure}

\minisection{Generating Deep Squish Pattern.} Once the training phase ends, we can synthesize fresh topology by sampling noise topology $\vec{T}_K$ from the stationary distribution, a random uniform distribution, and then gradually removing the predicted noise from it in the reverse procedure. The sampling process can be expressed by,
\begin{equation}
    p_\theta(\hat{\vec{T}}_0|\vec{T}_K) =p_\theta(\hat{\vec{T}}_{0}|\vec{T}_1) \prod_{k=2}^{K}p_\theta(\vec{T}_{k-1}|\vec{T}_k),
\end{equation}
where $\vec{T}_k$ is the estimated pattern topology at step $k$ and $\hat{\vec{T}}_0$ is the newly sampled topology tensor. The denoising procedure is illustrated in \Cref{fig:sampling}. The generated topology tensor $\hat{\vec{T}}_0$ is naturally a binary one where each entry equals either zero or one. When the sampling ends, we flatten the $\hat{\vec{T}}_0$ for Legal Pattern Assessment.

% After training, our model can generate new deep squish patterns without any real pattern input. We only need to sample $\vec{T}_K$ from the stationary distribution set by the model, and

\minisection{Topology Pre-filter.} We conduct a rule-based topology pre-processing here to filter out invalid topology, {\it e.g.,} Bow-tie, according to domain knowledge. Thanks to the high-quality topologies generated by our discrete diffusion model, only less than 0.1\% of generated topologies are filter-out by the pre-filter in our settings.

\subsection{2D Legal Pattern Assessment}
\label{sec:2.4}

Once the squish pattern generation is finished, we need the legal $\Delta_x$s and $\Delta_y$s of all generated topologies to generate DRC-clean layout patterns.
The decomposition of topology generation and legal pattern assessment brings flexibility to \tool{DiffPattern} when the design rules change, as we further discussed in \Cref{sec:4.3}.
Instead of using a black-box deep-learning-based method as previous works \cite{zhang2020layout,wen2022layoutransformer} do, we use a white-box method to solve the problem. We first list all constraints for each generated topology according to the design rules introduced in \Cref{fig:drc_rule} and then formulate a nonlinear system combining all of them, as in Formula (\ref{eq:nonlinear}), 
\begin{equation}
    \label{eq:nonlinear}
    \begin{cases}
        \delta_{xi}, \delta_{yj} > 0,                                                          &\forall \delta_{xi}, \delta_{yj};\\
        \sum \delta_{xi}  =\sqrt{C}M, \quad \sum \delta_{yj}  =\sqrt{C}M;                                       \\
        \sum_{i=a}^b \delta_{i} \geq\textit{Space}_\textit{min},                               &\forall (a,b)\in Set_{S};\\
        \sum_{i=a}^b \delta_{i} \geq\textit{Width}_\textit{min},                               &\forall (a,b)\in Set_W;\\
        \sum \delta_{xi}\delta_{yj}\in[\textit{Area}_\textit{min},\textit{Area}_\textit{max}], &\forall \text{ Polygon};
    \end{cases}
\end{equation}
where $\textit{Space}_\textit{min}$ and $\textit{Width}_\textit{min}$ are the lower bound of `Space' and `Width'. $\sqrt{C}M\times \sqrt{C}M$ is the shape of topology matrix.
$\textit{Area}_\textit{min}$ and $\textit{Area}_\textit{max}$ define the legal area range of each polygon in the pattern.
All the constants are pattern-independent and given by design rules.
Both $Set_{S}$ and $Set_W$ are pattern-dependent and indicate which pair of scan lines is constrained by design rules on `Space' and `Width', respectively. 

Note that the nonlinear system in \Cref{eq:nonlinear} can be efficiently solved with vast nonlinear programming algorithms or numerical methods, and the solution is usually not unique. Every solution of $\Delta_x$ and $\Delta_y$ together with the associated topology formulates a complete squish pattern representation.
As further detailed in \Cref{sec:4.3}, \tool{DiffPattern} can easily synthesize a large number of legal layout patterns from a single topology with the given design rules. Theoretically, there are cases where no legal solution is figured out in a limited time. Although it never happens in our experiments (more than $1.0\times 10^8$ times attempts), we can simply remove these unsolvable cases from the generated topology set to avoid synthesizing illegal patterns. 
In most cases, the choice of the initial value for the nonlinear programming algorithms may have a minor impact on pattern diversity and legality. In practice, we randomly choose a pair of existing geometric vectors from datasets as the start point, which will empirically accelerate the convergence of nonlinear programming algorithm.

\section{Experimental Results}

\begin{table*}[tb!]
    \centering
    \caption{Comparison on pattern diversity and legality. All results are copied from \cite{wen2022layoutransformer} and \cite{zhang2020layout}. `Real Patterns' refer to the whole dataset (Training set + Test set). `-' refer to Not Applicable.}
    \label{tab:diversity}
    %{{{
    %\resizebox{.78\linewidth}{!}
    {
        \begin{tabular}{c|c|cc|cc}
            \toprule
            \multirow{2}*{Set/Method}&\multirow{2}*{Generated Topology}&\multicolumn{2}{c|}{Generated Patterns}&\multicolumn{2}{c}{Legal Patterns}\\
            &&Patterns & Diversity ($\uparrow$)  &Legality ($\uparrow$) & Diversity ($\uparrow$)\\ \midrule
            \rowcolor{green!10} Real Patterns&-&-&-&13869&10.777\\
            CAE\cite{yang2019deepattern}&100000&100000&4.5875&19&3.7871 \\
            VCAE\cite{zhang2020layout}&100000&100000&\textbf{10.9311}&2126&9.9775 \\
            CAE+LegalGAN\cite{zhang2020layout}&100000&100000&5.8465&3740&5.8142 \\
            VCAE+LegalGAN\cite{zhang2020layout}&100000&100000&9.8692&84510&9.8669 \\
            LayouTransformer\cite{wen2022layoutransformer}&-&100000&10.532&89726&10.527 \\
            \tool{DiffPattern}-S&100000&100000&10.815&\textbf{100000}&\textbf{10.815} \\
            \tool{DiffPattern}-L&100000&10000000&10.815&\textbf{10000000}&\textbf{10.815}\\
            % DiffPattern-L&&&&&\\
            \bottomrule 
        \end{tabular}
    }
    %}}}
\end{table*}

\subsection{Experimental Setup}

\minisection{Datasets.} We follow previous work \cite{zhang2020layout, wen2022layoutransformer} to obtain the dataset of small layout pattern
images with the size of $2048\times2048$ $nm^2$ by splitting a $400\times160$ $\mu m^2$ layout map from ICCAD contest 2014. The size of extracted topology tensor is fixed as 16$\times$32$\times$32 with $C=16$ in Deep Squish Pattern Representation. 3000 images are randomly choosed as test set, while others are used for training. 

\begin{figure}[tb!]
    \centering
    \subfloat[]{ \includegraphics[width=0.28\linewidth]{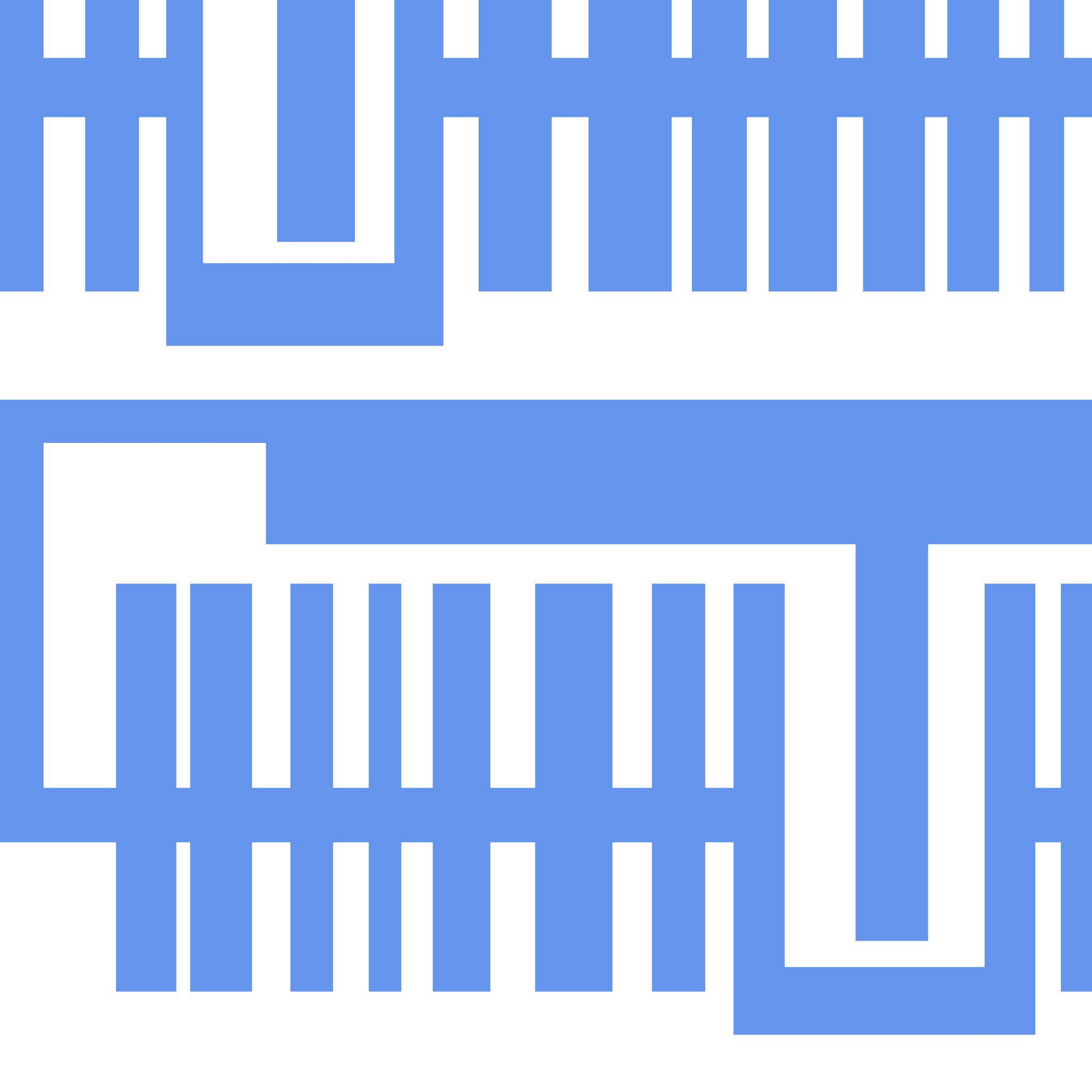} }
    \subfloat[]{ \includegraphics[width=0.28\linewidth]{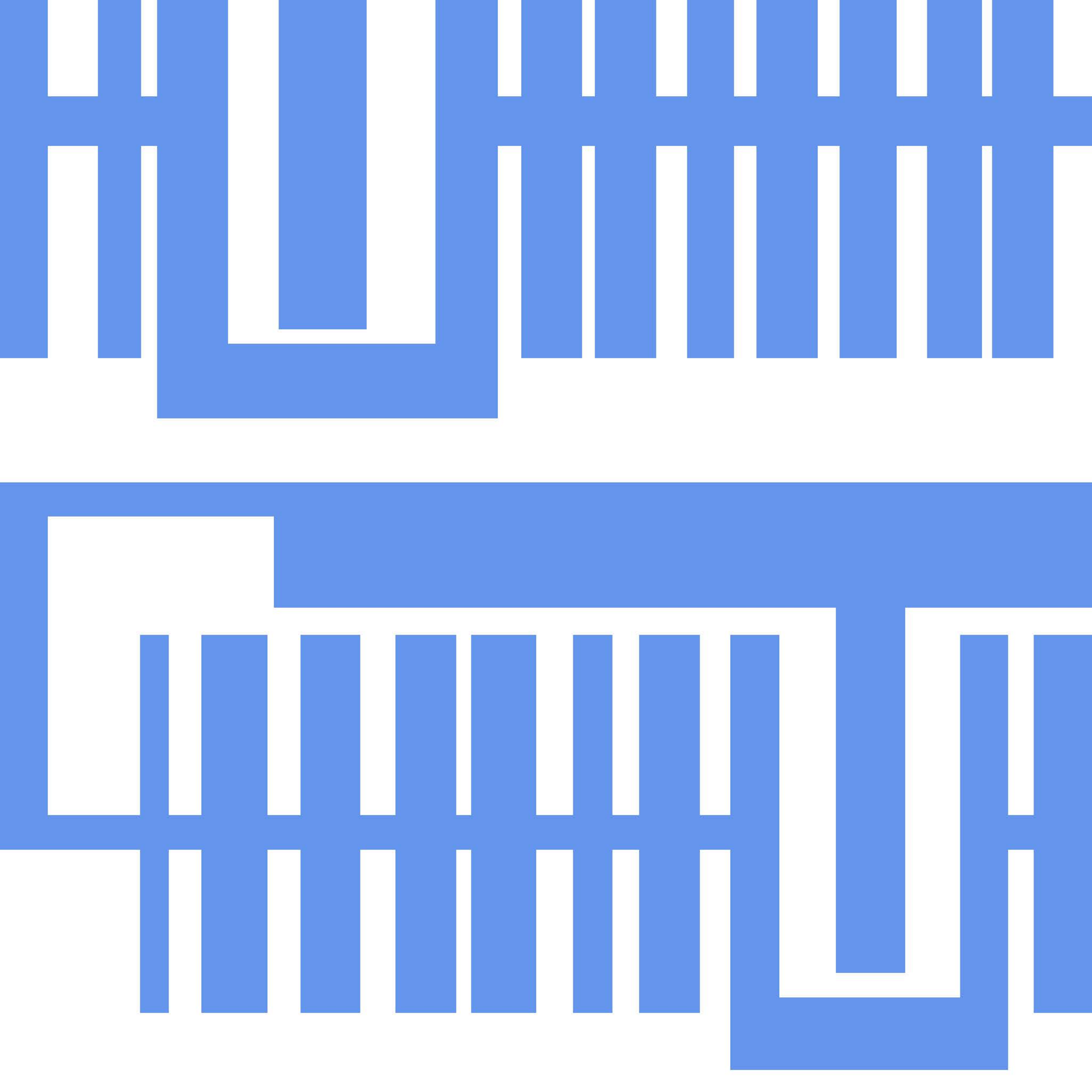} } 
    \subfloat[]{ \includegraphics[width=0.28\linewidth]{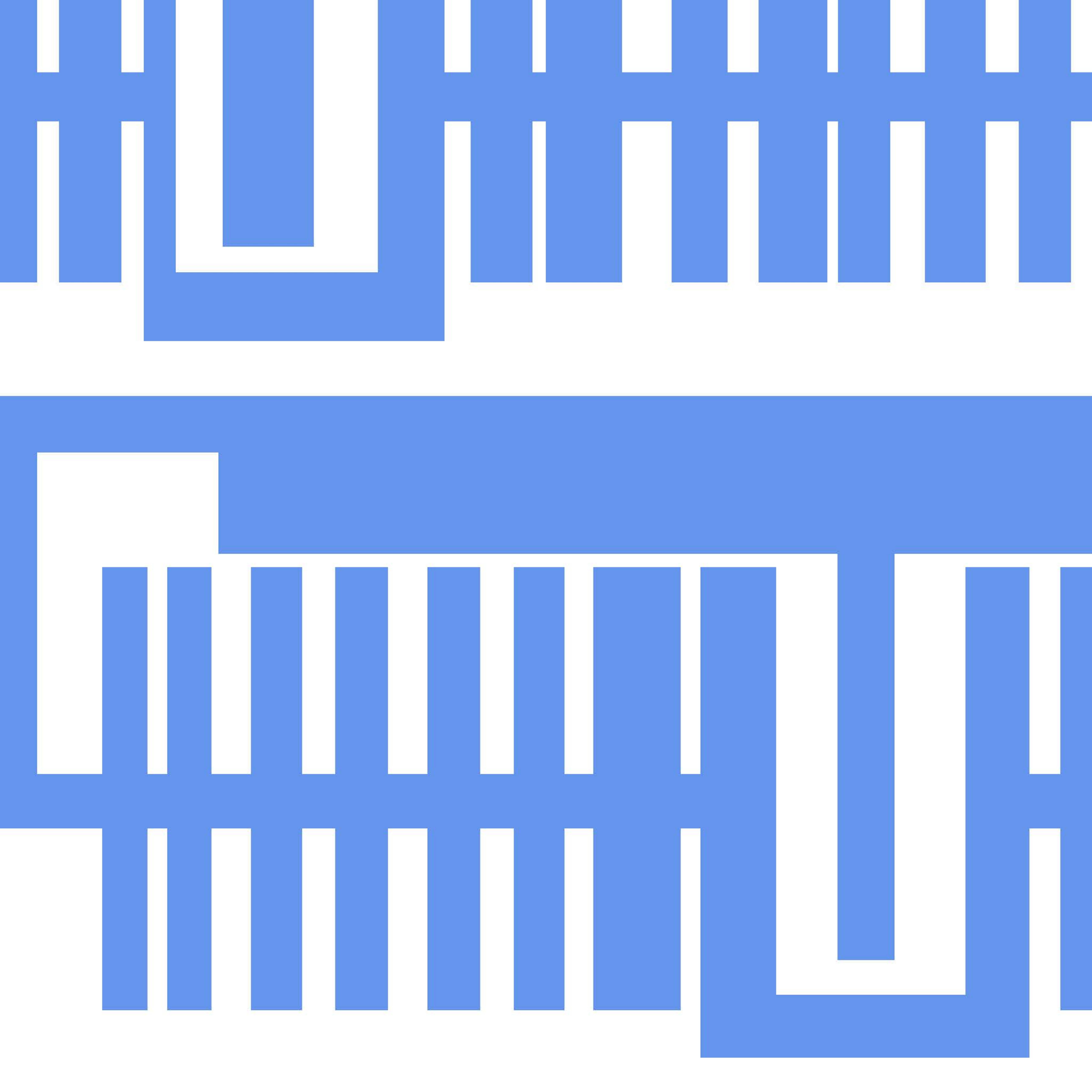} } \\
    \subfloat[]{ \includegraphics[width=0.28\linewidth]{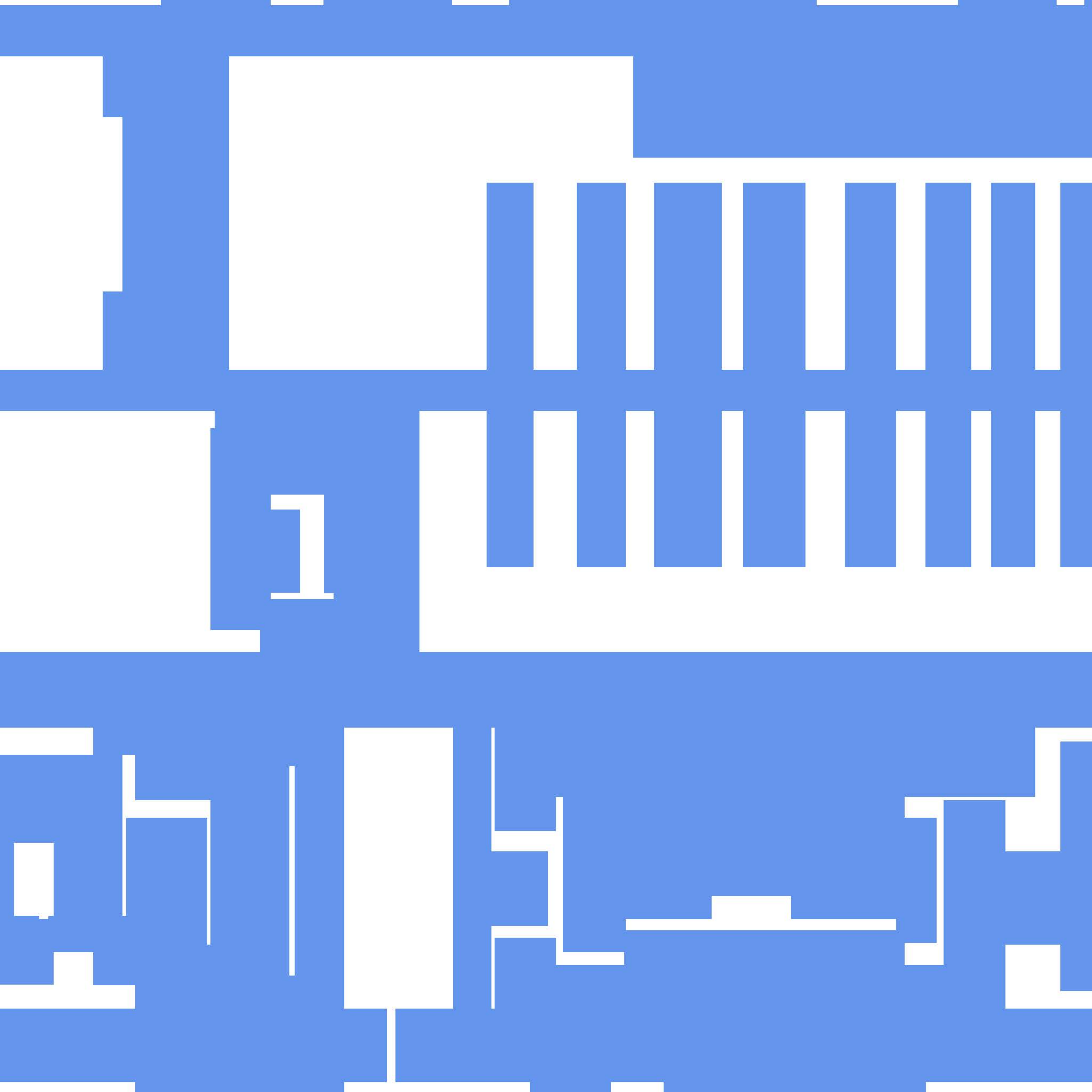} }
    \subfloat[]{ \includegraphics[width=0.28\linewidth]{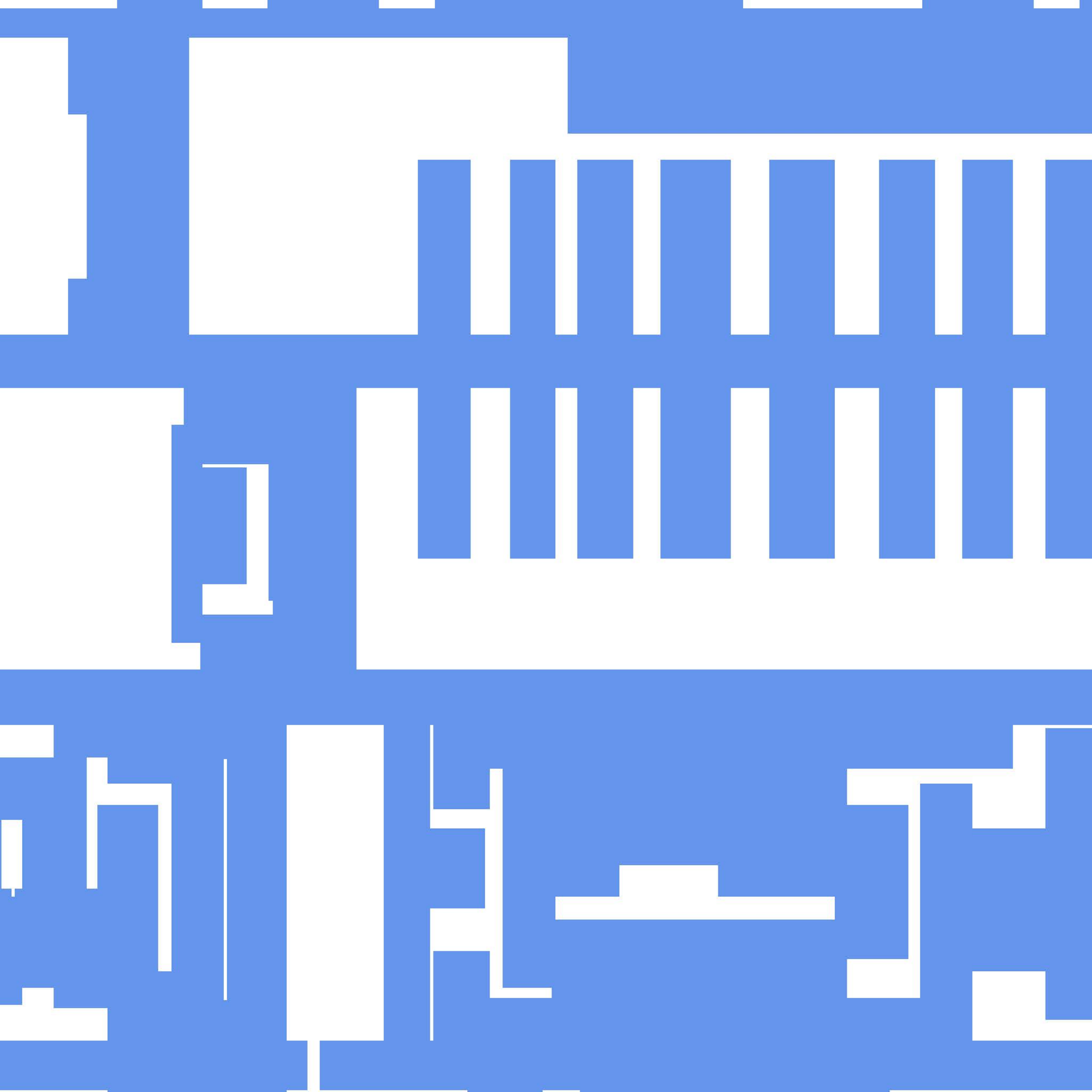} }
    \subfloat[]{ \includegraphics[width=0.28\linewidth]{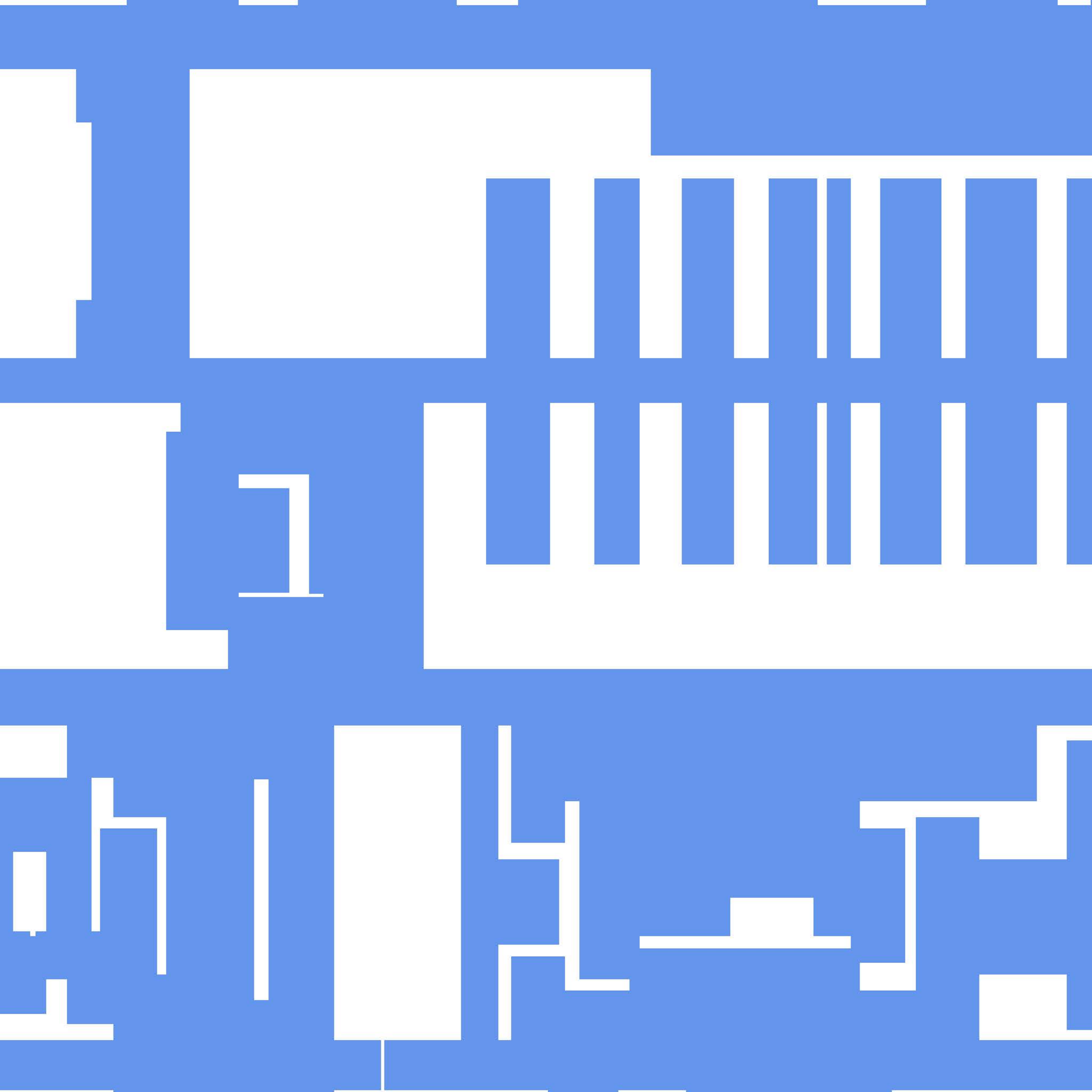} }
    \caption{Different layout patterns that are generated from a single topology with the same design rule. }
    \label{fig:single}
\end{figure}

\minisection{Diffusion Model Configuration.} Following the setting in previous works \cite{ho2020denoising,austin2021structured}, we use a U-Net \cite{ronneberger2015u} as our backbone in the discrete diffusion model to predict the posterior distribution during the reverse diffusion process. There are four feature map resolutions in the model, $[32\times32, 16\times16, 8\times8, 4\times4]$. Each resolution level has two convolutional residual blocks, where the numbers of convolution channels on the four resolutions are $\left[128,256,256,256\right]$, respectively. A self-attention block is also placed between two convolutional blocks at the 16$\times$16 resolution level. Moreover, the time step $k$ is included in each residual block through the sinusoidal position embedding \cite{vaswani2017attention}.

\minisection{Training Details.} We train the diffusion model for $0.5$M iterations with a batch size of $128$. The learning rate is $2\times 10^{-4}$. And we use the Adam optimizer. Other hyperparameters are chosen as following: the dropout rate is set to $0.1$, the grad clip is set to $1$, and the loss coefficient $\lambda$ is set to $0.001$. We set diffusion timesteps $K= 1000$ to ensure that the forward diffusion process converges to the uniform stationary distribution within $K$ steps. The noise schedule is: $\beta_k$ is linearly increased from $0.01$ to $0.5$. The training procedure takes about $17$ hours in total on 8 NVIDIA RTX 3090 GPUs.

\subsection{Pattern Diversity and Legality}
\label{sec:4.2}

The diversity is calculated by \Cref{eq:diveristy}, and the legality is checked by a tool {\it Klayout} based on the design rules described in \Cref{sec:2.3}. 

To have a fair comparison with previous methods, we randomly synthesize 100000 typologies. We denote a version of \tool{DiffPattern} as \tool{DiffPattern}-S where we only assign one pair of geometric vectors to each generated topology in the phase of legal pattern assessment.
However, \tool{DiffPattern} is able to generate a large amount of legal patterns for each topology.
To show the strong ability of our methodologies, we have another version denoted as \tool{DiffPattern}-L,
where we figure out one hundred different legal geometric vectors for each generated topology.
Considering the size of topology and the large search space, the number of legal solutions for each topology is huge. We `only' randomly take one hundred from them due to the time limitation.

We compare our \tool{DiffPattern} with several learning-based layout pattern generation methods.
CAE \cite{yang2019deepattern} denotes a vanilla convolutional auto-encoder model.
VCAE \cite{zhang2020layout} utilizes a variational convolutional auto-encoder model.
Both of them are pixel-based methods.
LegalGAN \cite{zhang2020layout} is a learning-based post-processing method that legalizes a newly generated topology by modifying it.
LayouTransformer \cite{wen2022layoutransformer} is a sequential-based method.
LayouTransformer applies a transformer-based model to synthesize new sequential representations of layout patterns without topology generation.
As shown in \Cref{tab:diversity}, thanks to the topology pre-filter and rule-based 2D legal pattern assessment,
both \tool{DiffPattern}-S and \tool{DiffPattern}-L achieve perfect performance ({\it i.e.}~100\%) under the metric of legality in the standard settings.
With the well-designed topology generation method, \tool{DiffPatten} also gets reasonable improvement (10.527$\rightarrow$10.815) on the diversity of generated pattern compared with the previous best method, LayouTransformer.
Furthermore, different from the previous methods, the legalization of \tool{DiffPattern} mainly relies on the rule-based legal pattern assessment and topology pre-filter.
When design rules change, it is easy for us to produce another batch of diverse patterns that satisfy the new design rules,
without re-training the topology generation model.
We detail the flexibility of our method in the next subsection.

\begin{figure}[tb!]
    \centering
    \subfloat[]{ \includegraphics[width=0.28\linewidth]{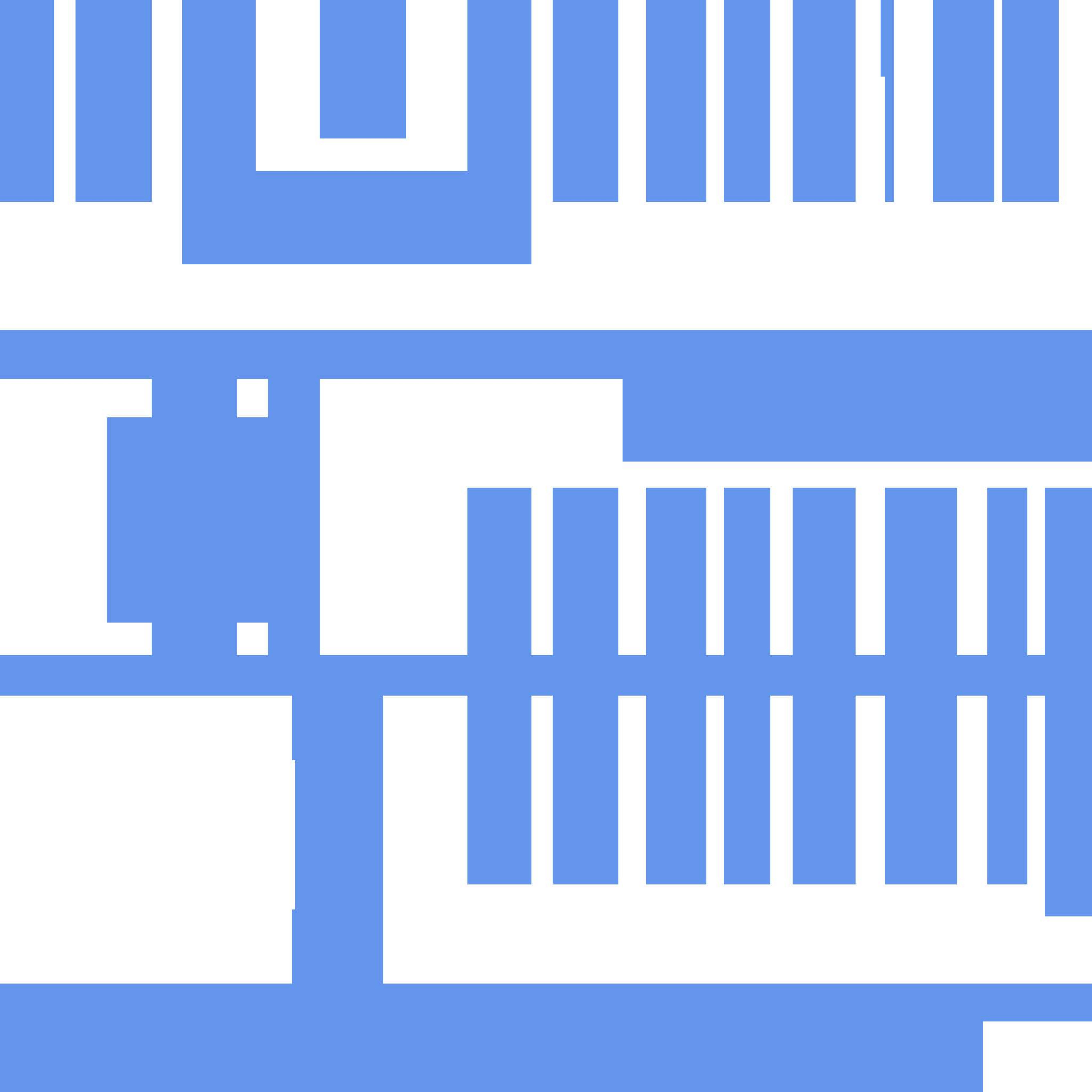} }
    \subfloat[]{ \includegraphics[width=0.28\linewidth]{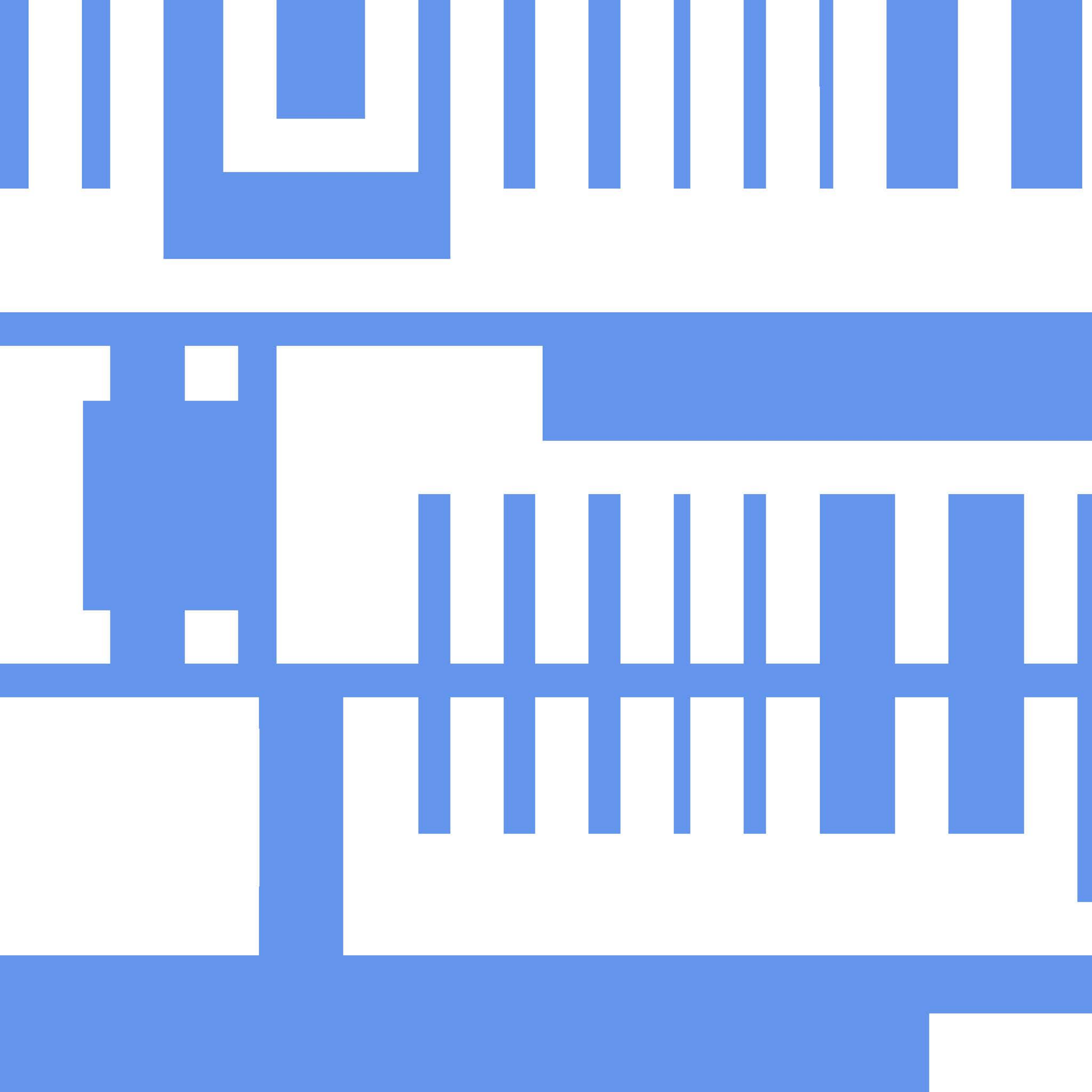} }
    \subfloat[]{ \includegraphics[width=0.28\linewidth]{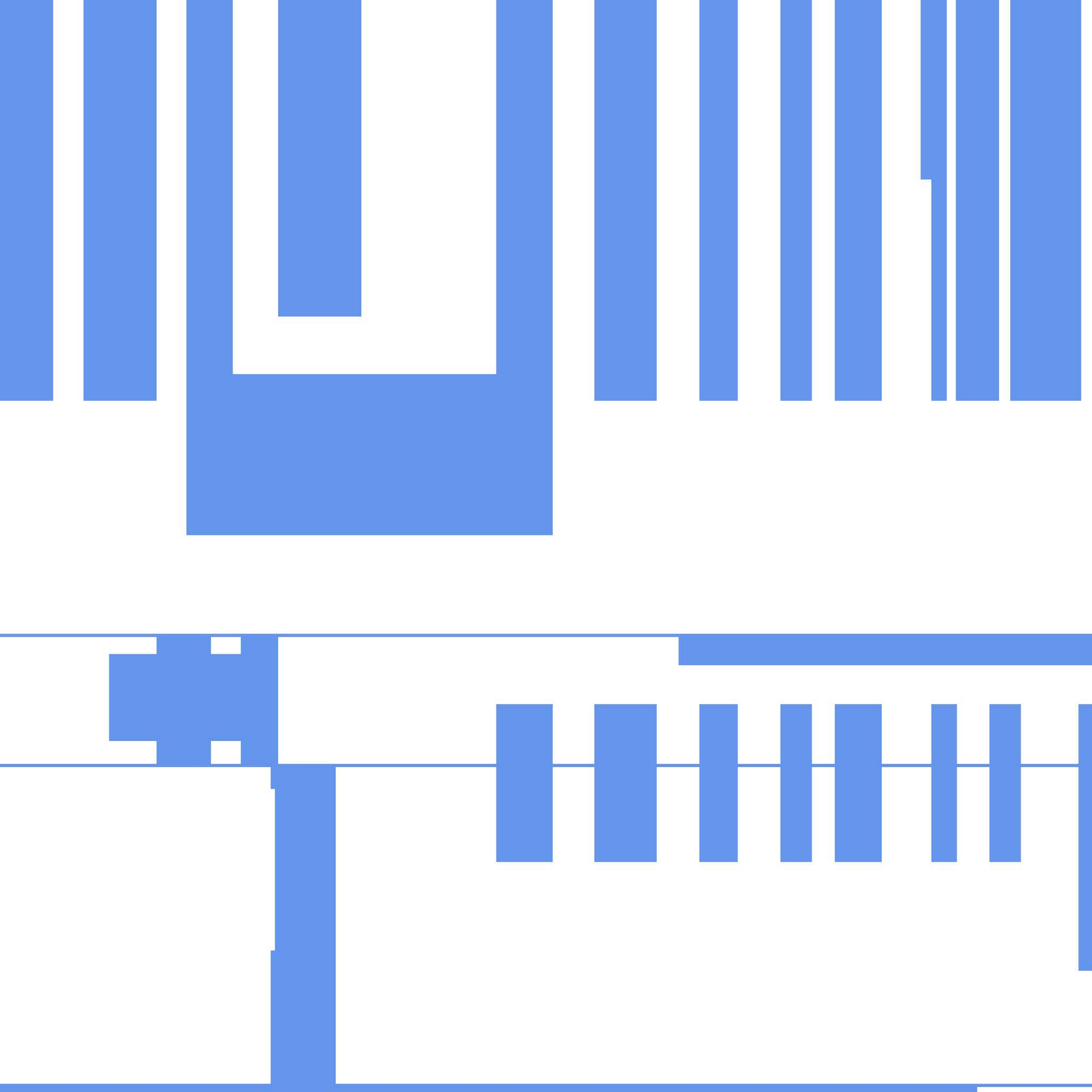} }
    \caption{
        Layout patterns that are generated from the same topology with different design rules: 
        (a) Normal rule; (b) Larger $\textit{space}_\textit{min}$; (c) Smaller $\textit{Area}_\textit{max}$.
    }
    \label{fig:rules}
\end{figure}

\subsection{Flexibility}
\label{sec:4.3}

A main advantage of \tool{DiffPattern} is that the white-box 2D legal pattern assessment phase is quite flexible. We show two applications of this property here. 

\minisection{Generate Different Patterns from Single Topology.} Given the design rules and the topology, the non-linear system in \Cref{eq:nonlinear} usually has many legal solutions, {\it i.e.,} the geometric vector pairs. Every legal solution leads to a legal layout pattern, and these layout patterns share a common topology, which are useful in some downstream tasks. We show the examples where several layout patterns are generated from a single topology with different geometric vectors in \Cref{fig:single}.

\minisection{Generate Legal Patterns with Different Design Rules.}
The legalization and topology generation are decoupled in \tool{DiffPattern},
which allows us to generate legal layout patterns under different design rules without re-training the model. We show the examples where several layout patterns are generated from the same topology but with different design rules in \Cref{fig:rules}.

\subsection{Distribution of Complexity}

\begin{figure}[tb!]
    \centering
    \includegraphics[width=\linewidth]{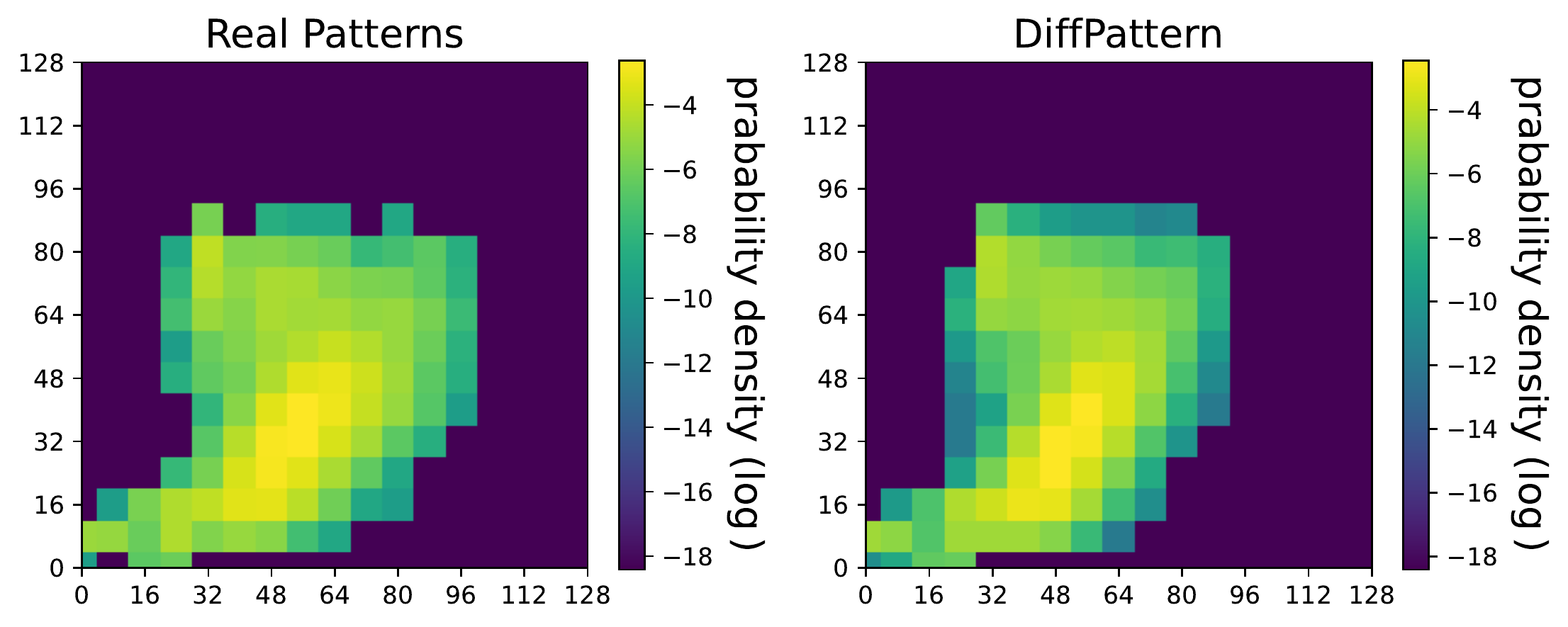}
    \caption{An illustration of complexity distribution. }
    \label{fig:complexity}
\end{figure}

Diversity is a critical metric of the quality of generated pattern library.
As defined in \Cref{sec:2.3}, diversity is the Shannon entropy of the distribution of pattern complexity,
{\it i.e.,} the number of scan lines subtracted by a pattern along the x-axis and y-axis.
We visualize the distribution of complexity in \Cref{fig:complexity}.
The patterns generated by \tool{DiffPattern} share a similar complexity distribution with real patterns.
This visualization further demonstrates our ability to generate high-quality layout patterns.

\subsection{Model Efficiency}

Model efficiency is an essential statistic for layout generation methods.
Since the topology generation and layout pattern assessment are decomposed in our model,
we record the average time to sample a new topology and figure out a legal solution for \Cref{eq:nonlinear} separately with our implementation.
The results are listed in \Cref{tab:efficiency}.
We have mentioned in \Cref{sec:2.4} that we randomly select a pair of existing geometric vectors to initialize the non-linear system for acceleration. We denote the version as {\it Solving-E}. We denote the original version with random initialization as {\it Solving-R}. Comparing with Solving-R, the proposed Solving-E version can achieve an average of $2.30\times$ acceleration.

\subsection{Discussion on Validity}

There is a metric named pattern validity proposed by previous work \cite{zhang2020layout}. The evaluation of validity is based on an encoder-decoder model that is pre-trained on the training set. The basic idea is if the generated patterns share more similar features with patterns in the training set, these new patterns will get a better score from the validity metric. However, we argue that the meaning of the metric of validity is quite limited. One main motivation of the layout pattern generation task is to synthesize a large amount of diverse but legal layout patterns for downstream tasks, {\it e.g.,} hotpot detection or lithography simulation. In these situations, legal patterns that are dissimilar from existing patterns are preferred but they get worse scores from validity. What makes the metric worse is that the measurement of validity encourages the overfitting of the training set. For example, according to \cite{zhang2020layout,wen2022layoutransformer}, the generated patterns even receive a much higher score (65\%$\rightarrow$84\%) than the patterns from the test set, which theoretically follows the same distribution as the training set.
It is unreasonable to believe the method with the better score in validity is superior in quality.
Therefore, we have not evaluated \tool{DiffPattern} with this metric. 

% The metric of pattern validity is proposed in \cite{zhang2020layout}, which aims to measure how realistic the generated patterns are. Furthermore, .  and are treated as the better ones. The validity counts the number of DRC-clean patterns whose scores are lower than the threshold $\tau$. 

\begin{table}[tb!]
    \centering
    \caption{Model efficiency of \tool{DiffPattern}. We record the average processing time for each sample (topology or layout pattern) in our machine. One Nvidia RTX 3090 GPU is used for topology sampling and one Intel(R) Xeon(R) Gold 6326 CPU @ 2.90GHz is used to figure out the non-linear system in this table. }
    \label{tab:efficiency}
    % \resizebox{.85\linewidth}{!}
    % {
        \begin{tabular}{c|cc}
            \toprule
            Phase/Method & Cost Time (s)& Acceleration  \\
            \midrule
            Sampling &0.544 &N/A\\
            \midrule
            Solving-R & 0.269&1.00$\times$\\
            Solving-E & 0.117&2.30$\times$\\
            
            \bottomrule 
        \end{tabular}
    % }
\end{table}

% 

% \begin{table}[tb!]
%     \centering
%     \caption{Comparison on pattern validity. All results are copied from \cite{wen2022layoutransformer} and \cite{zhang2020layout}}
%     \label{tab:diversity}
%     % \resizebox{.68\linewidth}{!}
%     {
%         \begin{tabular}{c|c|ccc}
%             \toprule
%             \multirow{2}*{Set/Method}&\multirow{2}*{Legal Patterns}&\multicolumn{3}{c}{Patten Validity($\uparrow$)}\\
%             && $\tau$=0.6  &$\tau$=0.7 & $\tau$=0.8 \\ \midrule
%             \rowcolor{green!10} Test Set&3000&0.6493&0.8485&0.9306\\
%             CAE+LegalGAN\cite{zhang2020layout}&3740&0.0003&0.0027&0.0167 \\
%             VCAE+LegalGAN\cite{zhang2020layout}&84510&0.5430&0.7840&0.9057 \\
%             LayouTransformer\cite{wen2022layoutransformer}&89726&\textbf{0.8416}&\textbf{0.9438}&\textbf{0.9834} \\
%             DiffPattern-R&100000&0.2863&0.6022&0.8436 \\
%             DiffPattern-E&100000&0.6016&0.8090&0.9234 \\
%             \bottomrule 
%         \end{tabular}
%     }
% \label{tab:validity}
% \end{table}

\section{Conclusion}
 We aim to synthesize diverse and legal layout patterns, and propose a practical method named \tool{DiffPattern}. Based on the given design rules our approach allows us to stably generate theoretically infinite legal layout patterns. The experiment results show that \tool{DiffPattern} outperforms previous SOTA methods by a large margin. Meanwhile, \tool{DiffPattern} is easy to extend and can be used on more complex pattern generation scenarios. In future work, we hope to extend our method to a more generic layout pattern generation approach.

 %We aim to transfer knowledge in the pose domain and propose an effective method named FlexPose. Our approach allows us to adapt an existing pose distribution to a different target one by using a few poses from the target dataset and generating theoretically infinite poses following the target distribution. FlexPose can be used on several pose-related works. In future work, we hope to extend our method to a more generic pose domain adaptation approach.

%\newpage
{
\bibliographystyle{IEEEtran}
\bibliography{ref/Top-sim,ref/FPGA-DNN,ref/FPGA,ref/FPGAPlacement,ref/tools,ref/DiffPattern}
}

\end{document}